\documentclass{article}

     \PassOptionsToPackage{numbers, compress}{natbib}

 \usepackage[final]{neurips_2020}

\usepackage[utf8]{inputenc} 
\usepackage[T1]{fontenc}    
\usepackage{hyperref}       
\usepackage{url}            
\usepackage{booktabs}       
\usepackage{amsfonts}       
\usepackage{nicefrac}       
\usepackage{microtype}      
\usepackage{amsthm}
\usepackage{amsmath}
\usepackage{enumitem}
 \usepackage[skip=0.5\baselineskip]{caption}
 \usepackage{subcaption}
\newtheorem{proposition}{Proposition}
\newtheorem{theorem}{Theorem}
\usepackage[]{mycomments}
\usepackage{comment}
\usepackage{tabularx}
\usepackage{natbib}
\usepackage{graphicx}
\usepackage{caption} 
\usepackage[ruled,vlined]{algorithm2e}

\title{Reservoir Computing meets Recurrent Kernels and Structured Transforms}

\author{Jonathan Dong\thanks{Equal contribution. Corresponding authors \texttt{\{jonathan.dong, 
ruben.ohana\}@ens.fr} } $^{\,1,2}$ \hspace{0.4cm} Ruben Ohana$^{*1,3}$ \hspace{0.4cm} Mushegh Rafayelyan$^2$ \hspace{0.4cm} Florent Krzakala$^{1,3,4}$\vspace{0.1cm}\\
$^1$Laboratoire de Physique de l’Ecole Normale Supérieure, ENS, Université PSL, CNRS, \\
Sorbonne Université, Université de Paris, F-75005 Paris, France \\
$^2$Laboratoire Kastler Brossel, Ecole Normale Supérieure, Université PSL, CNRS, \\
Sorbonne Université, Collège de France, F-75005 Paris, France \\ $^3$LightOn, F-75002 Paris, France\\ $^4$IdePHICS lab, Ecole Polytechnique Fédérale de Lausanne, Switzerland
}
\begin{document}

\maketitle
    
\begin{abstract}
Reservoir Computing is a class of simple yet efficient Recurrent Neural Networks where internal weights are fixed at random and only a linear output layer is trained. In the large size limit, such random neural networks have a deep connection with kernel methods. Our contributions are threefold: a) We rigorously establish the recurrent kernel limit of Reservoir Computing and prove its convergence. b) We test our models on chaotic time series prediction, a classic but challenging benchmark in Reservoir Computing, and show how the Recurrent Kernel is competitive and computationally efficient when the number of data points remains moderate. c) When the number of samples is too large, we leverage the success of structured Random Features for kernel approximation by introducing Structured Reservoir Computing. The two proposed methods, Recurrent Kernel and Structured Reservoir Computing, turn out to be much faster and more memory-efficient than conventional Reservoir Computing. 
\end{abstract}
\section{Introduction}
Understanding Neural networks in general, and how to train Recurrent Neural Networks (RNNs) in particular, remains a central question in modern machine learning. 
Indeed, backpropagation in recurrent architectures faces the problem of exploding or vanishing gradients \cite{pascanu2013difficulty, salehinejad2017recent}, reducing the effectiveness of gradient-based optimization algorithms. While there exist very powerful and complex RNNs for modern machine learning tasks, interesting questions still remain regarding simpler ones. In particular, Reservoir Computing (RC) is a class of simple but efficient Recurrent Neural Networks introduced in \cite{jaeger2001echo} with the Echo-State Network, where internal weights are fixed randomly and only a last linear layer is trained \cite{verstraeten2007experimental}. As the training reduces to a well-understood linear regression, Reservoir Computing enables us to investigate separately the complexity of neuron activations in RNNs. With a few hyperparameters, we can tune the dynamics of the reservoir from stable to chaotic and performances are increased
when RC operates close to the chaotic regime \cite{lukovsevivcius2009reservoir}.

Despite its simplicity, Reservoir Computing is not fully efficient: computational and memory costs grow quadratically with the number of neurons. 
To tackle this issue, efficient computation schemes have been proposed based on sparse weight matrices \cite{lukovsevivcius2009reservoir}. 
Moreover, there is an active community developing novel hardware solutions for energy-efficient, low-latency RC \cite{pathak2018model}. Based on dedicated electronics \cite{antonik2015fpga, wang2015general, zhang2015digital, jin2017performance}, optical computing \cite{larger2012photonic, duport2012all, van2017advances, dong2018scaling, dong2019optical}, or other original physical designs \cite{tanaka2019recent}, they leverage the robustness and flexibility of RC. 
Reservoir Computing has already been used in a variety of tasks, such as speech recognition and robotics \cite{lukovsevivcius2012reservoir} but also combined with Random Convolutional Neural Networks for image recognition~\cite{tong2018reservoir} and Reinforcement Learning~\cite{chang2020reinforcement}. A very promising application today is chaotic time series prediction, where the RC dynamics close to chaos may prove a very important asset \cite{pathak2018model}.
Reservoir Computing also represents an important model in computational neuroscience, as parallels can be drawn with specific regions of the brain behaving like a set of randomly-connected neurons \cite{hinaut2013real}.

As RC embeds input data in a high-dimensional reservoir, it has already been linked with kernel methods \cite{lukovsevivcius2009reservoir}, but merely as an interesting interpretation for discussion. In our opinion, this point of view has not been exploited to its full potential yet. 
A study derived the explicit formula of the corresponding recurrent kernel associated with RC \cite{hermans2012recurrent}, this important result meaning the infinite-width limit of RC is a deterministic Recurrent Kernel (RK). 
Still, no theoretical study of convergence towards this limit has been conducted previously and the computational complexity of Recurrent Kernels has not been derived yet.

In this work, we draw the link between RC and the rich literature on Random Features for kernel approximation \cite{rahimi2008random, rahimi2009weighted, rudi2017generalization, carratino2018learning, liu2020random}. To accelerate and scale-up the computation of Random Features, one can use optical implementations~\cite{saade2016random, ohana2020kernel} or structured transforms \cite{le2013fastfood, yu2016orthogonal}, providing a very efficient method for kernel approximation. Structured transforms such as the Fourier or Hadamard transforms can be computed in $O(n \log n)$ complexity and, coupled with random diagonal matrices, they can replace the dense random matrix used in Random Features. 

Finally, we note that Reservoir Computing can be unrolled through time and interpreted as a multilayer perceptron. The theoretical study of such randomized neural networks through the lens of kernel methods has attracted a lot of attention recently \cite{jacot2018neural,mei2019generalization,gallicchio2020deep}, which provides a further motivation to our work. Some parallels were already drawn between Recurrent Neural Networks and kernel methods \cite{liang2019kernelRNNs, chen_mairal2019recurrent}, but they do not tackle the high-dimensional random case of Reservoir Computing. 

\paragraph{Main contributions ---} Our goal in this paper is to bridge the gap between the considerable amount of results on kernels methods, random features --- structured or not --- and Reservoir Computing. 

First, we rigorously prove the convergence of Reservoir Computing towards Recurrent Kernels provided standard assumptions and derive convergence rates in $O(1/\sqrt N)$, with $N$ being the number of neurons. We then numerically show convergence is achieved in a large variety of cases and does not occur in practice only when the activation function is unbounded (for instance with ReLU). 

When the number of training points is large, the complexity of RK grows; this is a common drawback of kernel methods. To circumvent this issue, we propose to accelerate conventional Reservoir Computing by replacing the dense random weight matrix with a structured transform. In practice, Structured Reservoir Computing (SRC) allows to scale to very large reservoir sizes easily, as it is faster and more memory-efficient than conventional Reservoir Computing, without compromising performance. 

These techniques are tested on chaotic time series prediction, and they all present comparable results in the large-dimensional setting. We also derive the computational complexities of each algorithm and detail how Recurrent Kernels can be implemented efficiently. 
In the end, the two acceleration techniques we propose are faster than Reservoir Computing and can tackle equally complex tasks. A public repository is available at \url{https://github.com/rubenohana/Reservoir-computing-kernels}.
\section{Recurrent Kernels and Structured Reservoir Computing}

Here, we briefly describe the main concepts used in this paper. We recall the definition of Reservoir Computing and Random Features, define Recurrent Kernels (RKs) and introduce Structured Reservoir Computing (SRC).

\textbf{Reservoir Computing (RC)} as a Recurrent Neural Network receives a sequential input $i^{(t)} \in \mathbb{R}^d$, for $t \in \mathbb{N}$. We denote by $x^{(t)} \in \mathbb{R}^N$ the state of the reservoir, $N$ being the number of neurons in the reservoir. Its dynamics is given by the following recurrent equation:
\begin{equation}
	\label{eq: initial recurrent equation}
	x^{(t+1)} = \frac{1}{\sqrt N} f\left(W_r\,x^{(t)} + W_i\,i^{(t)}\right)
\end{equation}
where $W_r \in \mathbb{R}^{N \times N}$ and $W_i \in \mathbb{R}^{N \times d}$ are respectively the reservoir and input weight matrices. They are fixed and random: each weight is drawn according to an i.i.d. gaussian distribution with variances $\sigma_r^2$ and $\sigma_i^2$, respectively. Finally, $f$ is an element-wise non-linearity, typically a hyperbolic tangent. To refine the control of the reservoir dynamics, it is possible to add a random bias and a leak rate. In the following, we will keep the minimal formalism of Eq. (\ref{eq: initial recurrent equation}) for  conciseness. 

We use the reservoir to learn how to predict a given output $o^{(t)} \in \mathbb{R}^c$ for example. The output predicted by the network $\hat{o}^{(t)}$ is obtained after a final layer:
\begin{equation}
	\label{eq: linear output layer equation}
	\hat{o}^{(t)} = W_o\, x^{(t)}
\end{equation}
Since only these output weights $W_o \in \mathbb{R}^{c \times N}$ are trained, the optimization problem boils down to linear regression. Training is typically not a limiting factor in RC, in sharp contrast with other neural network architectures. The expressivity and power of Reservoir Computing rather lies in the high-dimensional non-linear dynamics of the reservoir. 

\textbf{Kernel methods} are non-parametric approaches to learning. Essentially, a kernel is a function measuring a correlation between two points $u,v \in\mathbb{R}^p$. A specificity of kernels is that they can be expressed as the inner product of feature maps $\varphi:\mathbb{R}^p\rightarrow\mathcal{H}$ in a possibly infinite-dimensional Hilbert space $\mathcal{H}$, i.e. $k(u,v)=\langle \varphi(u),\varphi(v)\rangle_\mathcal{H}$.
Kernel methods enable the use of linear methods in the non-linear feature space $\mathcal{H}$. Famous examples of kernel functions are the Gaussian kernel $k(u,v)=\exp\left(-\frac{\|u-v\|^2}{2\sigma^2}\right)$ or the arcsine kernel $k(u,v)=\frac2\pi \arcsin{\frac{\langle u,v\rangle}{\|u\|\|v\|}}$. When the dataset becomes large, it is expensive to numerically compute the kernels between all pairs of data points. 

\textbf{Random Features} have been developed in  \cite{rahimi2008random} to overcome this issue. This celebrated technique introduces a random mapping $\phi:\mathbb{R}^p\rightarrow\mathbb{R}^N$ such that the kernel is approximated in expectation:
\begin{equation}
\label{eq: kernel approx definition}
    k(u,v) = \left\langle \varphi(u),\varphi(v) \right\rangle_\mathcal{H} \approx \left\langle \phi(u), \phi(v) \right\rangle_{\mathbb{R}^N}
\end{equation}
with $\phi(u)= \frac{1}{\sqrt{N}}[f(\langle w_1,u\rangle),...,f(\langle w_N,u\rangle)]^\top\in\mathbb{R}^N$ and random vectors $w_1,...,w_N\in\mathbb{R}^p$. Depending on 
$f$ and the distribution of $\{w_i\}_{i=1}^N$, we can approximate different kernel functions. 

There are two major classes of kernel functions: translation-invariant \textbf{(TI)} kernels and rotation-invariant \textbf{(RI)} kernels. In our study, we will consider TI kernels of the form $k(u,v) = k(\|u-v\|_2^2)$ and RI kernels of the form $k(u, v) = k(\langle u, v\rangle )$. Both can be approximated using Random Features  \cite{rahimi2008random,kar2012random}. For example, Random Fourier Features (RFFs) defined by:
\begin{equation}
    \label{eq: TI RF map}
    \phi(u) = \frac{1}{\sqrt{N}}[\cos(\langle w_1,u\rangle),..., \cos(\langle w_N,u\rangle), \sin(\langle w_1,u\rangle),...,\sin(\langle w_N,u\rangle)]^\top
\end{equation}
approximate any TI kernel (provided $k(0) = 1$).  For example, when $w_1,...,w_N \sim \mathcal{N}(0,\sigma^{-2}I_p)$, we approximate the Gaussian kernel. 
A detailed taxonomy of Random Features can be found in \cite{liao_couillet2018spectrum}. 

Random Features can be more computationally efficient than kernel methods, when their number $N$ is smaller than the number of data points $n$. For this particular reason, Random Features are a method of choice to implement large-scale kernel-based methods. 

\textbf{Link with Reservoir Computing.} It is straightforward to notice that reservoir iterations of Eq. (\ref{eq: initial recurrent equation}) can be interpreted as a Random Feature embedding of a vector $[x^{(t)}, i^{(t)}]$ (of dimension $p = N+d$), multiplied by $W = [W_r, W_i]$. This means the inner product between two reservoirs $x^{(t)}$, $y^{(t)}$ driven respectively by two inputs $i^{(t)}$ and $j^{(t)}$ converges to a deterministic kernel as $N$ tends to infinity:
\begin{equation}
    \label{eq: random feature convergence}
       \langle x^{(t+1)},y^{(t+1)}\rangle\approx k([x^{(t)}, i^{(t)}], [y^{(t)}, j^{(t)}])
\end{equation}
As explained previously, this kernel depends on the choice of $f$ and the distribution of $W_r$ and $W_i$. 

By denoting $l^{(t)}=\sigma_i^2\langle i^{(t)},j^{(t)}\rangle$ and $\Delta^{(t)}= \sigma_i^2\|i^{(t)}-j^{(t)}\|^2$, TI and RI kernels are then of the form:
\begin{align}
        k([x^{(t)}, i^{(t)}], [y^{(t)}, j^{(t)}]) &=  k(\sigma_r^2\langle x^{(t)},y^{(t)}\rangle+l^{(t)})\label{eq: DP kernels}\quad \textrm{(RI)} \\
        &= k(\sigma_r^2\|x^{(t)}-y^{(t)}\|^2+ \Delta^{(t)})
        \label{eq: TI kernels}\quad \textrm{(TI)}
\end{align}


\textbf{The Recurrent Kernel limit.} Looking at Eq. (\ref{eq: DP kernels}) and (\ref{eq: TI kernels}), we notice the kernel at time $t$ depends on approximations of kernels at previous times in a recursive manner. Here, we introduce Recurrent Kernels to remove the dependence in $x^{(t)}$ and $y^{(t)}$.

We suppose for the sake of simplicity $x^{(0)} = y^{(0)} = 0$. We define RI recurrent kernels as:
\begin{equation}
\label{eq: recurrence k_t RI}
    \begin{cases}
      k_1\left(l^{(0)}\right) = k\left(l^{(0)}\right) \\
      k_{t+1}\left(l^{(t)},...,l^{(0)}\right) = k\left(\sigma_r^2k_{t}\left(l^{(t-1)},...,l^{(0)}\right)+l^{(t)}\right), \quad \text{for } t \in \mathbb{N}^*
    \end{cases}       
\end{equation}
Similarly for TI recurrent kernels with Random Fourier Features, exploiting the property in Eq. (\ref{eq: TI RF map}) that $\|x^{(t)}\|^2=\|y^{(t)}\|^2=1$:
\begin{equation}
    \label{eq: recurrence k_t TI}
    \begin{cases}
      k_1\left(\Delta^{(0)}\right) = k\left(\Delta^{(0)}\right)\\
      k_{t+1}\left(\Delta^{(t)},...,\Delta^{(0)}\right) = k\left(\sigma_r^2\left(2 - 2 k_{t}\left(\Delta^{(t-1)},...,\Delta^{(0)}\right)\right)+\Delta^{(t)}\right), \quad \text{for } t \in \mathbb{N}^*
    \end{cases}       
\end{equation}

These Recurrent Kernel definitions describe hypothetical asymptotic limits of large-dimensional Reservoir Computing, interpreted as recurrent Random Features. We will study in Section \ref{subsec: cvg theorem} the convergence towards this limit. 

\textbf{Structured Reservoir Computing.} In the Random Features literature, it is common to use structured transforms to speed-up computations  of random matrix multiplications ~\cite{le2013fastfood, yu2016orthogonal}. They have also been introduced for trained architectures, with Deep \cite{moczulski2015acdc} and Recurrent Neural Networks \cite{arjovsky2016unitary}. 

We propose to replace the dense random weight matrices $W = [W_r, W_i]$ by a succession of Hadamard matrices $H$ (structured orthonormal matrices composed of $\pm 1/\sqrt{p}$ components) and diagonal random matrices $D_i$ for $i = 1, 2, 3$ sampled from an i.i.d. Rademacher distribution \cite{yu2016orthogonal}: 
\begin{equation}
    \label{eq: structured transform}
    W = \frac{\sqrt{p}}{\sigma}H D_1 H D_2 H D_3
\end{equation}
We use the Hadamard transform for its simplicity and the availability of high-performance libraries in ~\cite{thomas2018learning}. This structured transform provides the two main properties of a dense random matrix: mixing the activation of the neurons (Hadamard transform) and randomness (diagonal matrices). 

\section{Convergence theorem and computational complexity}
\subsection{Convergence rates} 
\label{subsec: cvg theorem}
Our first main result is a convergence theorem of Reservoir Computing to its kernel limit. We use Bernstein's concentration inequality in our recurrent setting. Several assumptions will be necessary:
\begin{itemize}[leftmargin=10pt]
\item The kernel function $k$ is Lipschitz-continuous with constant $L$, i.e. $|k(a)-k(b)|\leq L |a-b|$.
\item The random matrices $W_{r}$ and $W_{i}$ are resampled for each $t$ to obtain uncorrelated reservoir updates: 
$x^{(t+1)}= \frac{1}{\sqrt{N}}f(W_{r}^{(t)}x^{(t)}+W_{i}^{(t)} i^{(t)})$. This assumption is required for our theoretical proof of convergence, but we show convergence is reached numerically even without redrawing the weight matrices, which is standard in Reservoir Computing (in Fig. \ref{fig:lipschitz}). 
\item The function $f$ is bounded by a constant $\kappa$ almost surely, i.e. $|f(W_{res}^{(t)}x^{(t)} + W_{in}^{(t)}i^{(t)})|\leq \kappa$.
\end{itemize}
\vspace{0.1cm}
\begin{theorem}\label{theorem1: RI kernels}(Rotation-invariant kernels) 
For the RI recurrent kernel defined in Eq. (\ref{eq: recurrence k_t RI}),
under the assumptions detailed above, and with $\Lambda = \sigma_r^2 L$. For all $t\in\mathbb{N}$, the following inequality is satisfied for any $\delta>0$ with probability at least $1-2(t+1)\delta$:
\begin{align}
    \label{eq: main theorem bound}
    \left|\langle x^{(t+1)},y^{(t+1)}\rangle  - k_{t+1}(l^{(t)},...,l^{(0)})\right| 
    &\leq \frac{1-\Lambda^{t+1}}{1-\Lambda} \Theta(N) &&\textrm{if} \quad \Lambda\neq 1 \\
    \label{eq: main theorem bound 2}
    &\leq (t+1) \Theta(N) &&\textrm{if} \quad \Lambda=1
\end{align}
with $\Theta(N) = \frac{4\kappa^2\log\frac1\delta}{3N}+2\kappa^2\sqrt{\frac{2\log\frac1\delta}{N}}$.
\end{theorem}
\vspace{-0.2cm} 
\begin{proof}

We use the following Proposition (Theorem 3 of \cite{boucheron2003concentration} restated in Proposition 1 of  \cite{rudi2017generalization}):
\begin{proposition}
\label{Prop1: Bernstein}
    (Bernstein inequality for a sum of random variables). Let $X_1,...,X_N$ be a sequence of i.i.d. random variables on $\mathbb{R}$ with zero mean. If there exist $R,S\in\mathbb{R}$ such that $|X_i| \leq R$ almost everywhere and $\mathbb{E}[X_i^2]\leq S$ for $i\in\{1,...,N\}$, then for any $\delta>0$ the following holds with probability at least $1-2\delta$:
    \begin{equation}
        \left|\frac1N\sum_{i=1}^N X_i\right| \leq \frac{2R\log\frac1\delta}{3N}+\sqrt{\frac{2S\log\frac1\delta}{N}}
    \end{equation}
\end{proposition}
Under the assumptions, Proposition~\ref{Prop1: Bernstein} yields with probability greater than $1-2\delta$:
\begin{equation}
    \label{eq: bernstein scalar product}
    \left|\langle x^{(t+1)},y^{(t+1)}\rangle -k([x^{(t)}, i^{(t)}],[y^{(t)}, j^{(t)}])\right| \leq \frac{4\kappa^2\log\frac1\delta}{3N}+2\kappa^2\sqrt{\frac{2\log\frac1\delta}{N }} = \Theta(N)
\end{equation}
It means the larger the reservoir, the more Random Features $N$ we sample, and the more the inner product of reservoir states concentrates towards its expectation value, at a rate $O(1/\sqrt{N})$. We now apply this inequality recursively to complete the proof, based on the observation that both Eq. (\ref{eq: main theorem bound}) and (\ref{eq: main theorem bound 2}) are equivalent to: $\left|\langle x^{(t+1)},y^{(t+1)}\rangle  - k_{t+1}(l^{(t)},...,l^{(0)})\right| \leq (1 + \Lambda + \Lambda^2 +... + \Lambda^t)\Theta(N)$.

For $t = 0$, provided $x^{(0)} = y^{(0)} = 0$, we have, according to Eq.~\ref{eq: bernstein scalar product}, with probability at least $1 - 2 \delta$:
\begin{equation}
    \left|\langle x^{(1)}, y^{(1)}\rangle - k_1(l^{(0)})\right|\leq \Theta(N)
\end{equation}

For any time $t \in \mathbb{N}^*$, let us assume the following event $A_{t}$ is true with probability $\mathbb{P}(A_{t}) \geq 1 - 2 t \delta$:
\begin{equation}
    \left|\langle x^{(t)}, y^{(t)}\rangle - k_{t} (l^{(t-1)},...,l^{(0)}) \right|
    \leq (1 + \ldots + \Lambda^{t-1}) \Theta(N)
\end{equation}
Using the Lipschitz-continuity of $k$, this inequality is equivalent to:
\begin{equation}
    \label{eq: proof final eq to t}
    \left|
    k(\sigma_r^2 \langle x^{(t)}, y^{(t)}\rangle + l^{(t)})
    - k(\sigma_r^2 k_{t} (l^{(t-1)},...,l^{(0)}) + l^{(t)}) \right|
    \leq (\Lambda + \ldots + \Lambda^{t}) \Theta(N)
\end{equation}

With Eq. (\ref{eq: bernstein scalar product}), the following event $B_{t}$ is true with probability $\mathbb{P}(B_{t}) \geq 1 - 2 \delta$:
\begin{equation}
\label{eq: proof eq t+1}
\bigg|\langle x^{(t+1)}, y^{(t+1)}\rangle- k\big(\sigma_r^2\langle x^{(t)}, y^{(t)}\rangle+l^{(t)}\big)\bigg|\leq \Theta(N)    
\end{equation}

Summing Eq. (\ref{eq: proof final eq to t}) and (\ref{eq: proof eq t+1}), with the triangular inequality and a union bound, the following event $A_{t+1}$ is true with probability $\mathbb{P}(A_{t+1}) \geq \mathbb{P}(B_{t} \cap A_t) = \mathbb{P}(B_{t})+\mathbb{P}(A_{t})-\mathbb{P}(B_{t}\cup A_{t})\geq 1-2\delta + 1-2t\delta - 1 \geq 1 - 2 (t+1) \delta$:
\begin{equation}
    \left|\langle x^{(t+1)}, y^{(t+1)}\rangle
    -
    k_{t+1} (l^{(t)},...,l^{(0)})\right|
    \leq (1 + \ldots + \Lambda^{t}) \Theta(N)
\end{equation}
\end{proof}
A statement and proof of a similar convergence bound for TI recurrent kernels is provided in the Supplementary Material. 

\subsection{Numerical study of convergence}
The previous theoretical study required three important assumptions that may not be valid for Reservoir Computing in practice. Moreover, there is still no rigorous proof on the convergence of Structured Random Features in the non-recurrent case due to the difficulty to deal with correlations between them. Thus, we numerically investigate whether convergence of RC and SRC towards the Recurrent Kernel limit is achieved in practice.

\begin{figure}[t!]
    \centering
    \includegraphics[width=\linewidth ]{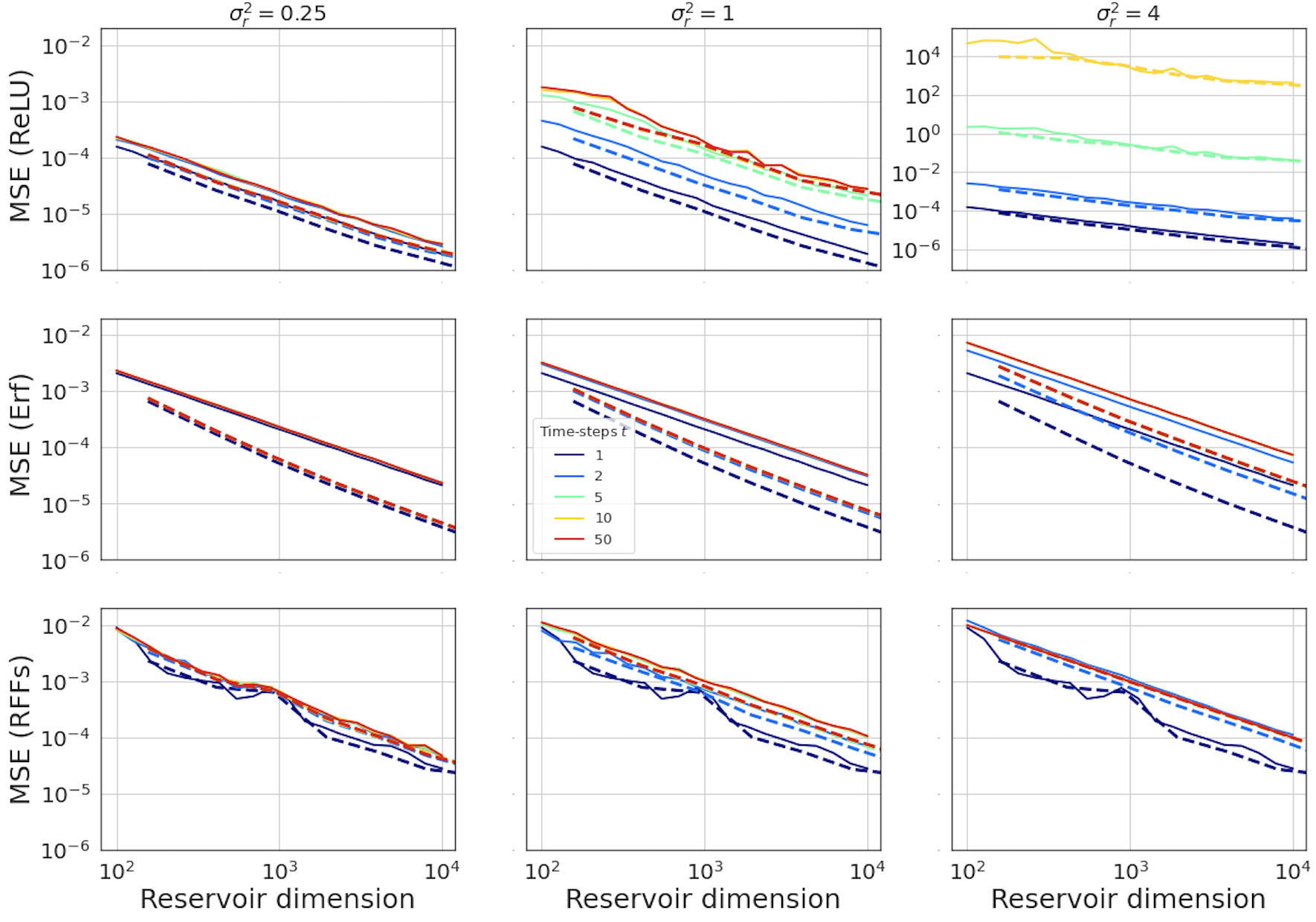}
    \caption{Convergence of Reservoir Computing towards its Recurrent Kernel limit for different variances of the reservoir weights $\sigma_r^2$ (columns), activation functions (lines: ReLU, Erf, RFFs) and times, for RC \textbf{(solid lines)} and SRC \textbf{(dashed lines)}. We observe that for the two bounded activation functions (Erf and RFFs), RC always converge towards the RK limit even at large times $t$. For ReLU, RC converges when $\sigma_r^2 = 0.25$ and $1$, and diverges as $t$ increases when $\sigma_r^2 = 4$. We also observe that SRC always yields equal or faster convergence than RC. The MSE decreases with an $O(1/N)$ scaling, which is consistent with the convergence rates derived in Theorem \ref{theorem1: RI kernels}.}
    \label{fig:lipschitz}
\end{figure}

In Fig.~\ref{fig:lipschitz}, we numerically compute the Mean-Squared Error (MSE) between the inner products obtained with a Recurrent Kernel and RC/SRC for different number of neurons in the reservoir. We generate 50 i.i.d. gaussian input time series $i^{(t)}_k$ of length $T$, for $k = 1, \ldots, 50$ and $t = 0, \ldots, T-1$. Each time series is fed into 50 reservoirs that share the same random weights, for RC and SRC. We compute the MSE between inner products of pairs of final reservoir states $\langle x^{(T)}_k, x^{(T)}_l \rangle$ and the deterministic limit obtained directly with $k_T(i^{(T-1)}_k, i^{(T-1)}_l, \ldots, i^{(0)}_k, i^{(0)}_l)$, for all $k, l = 1, \ldots, 50$. The computation is vectorized to be efficiently implemented on a GPU. Three different activation functions, the rectified linear unit (ReLU), the error function (Erf), and Random Fourier Features defined in Eq. \eqref{eq: TI RF map}, have been tested with different variances of the reservoir weights. The larger the reservoir weights, the more unstable the reservoir dynamics becomes.

Nonetheless, convergence is achieved in a large variety of settings, even when the assumptions of the previous theorem are not satisfied. For example, the ReLU non-linearity is not bounded and converges when $\sigma_r^2 \geq 1$. It is interesting to notice even for a large variance $\sigma_r^2 = 4$ do Reservoir Computing and Structured Reservoir Computing converge towards the RK limit for the second and third activation functions. This behavior has been consistently observed with any bounded $f$. 

On the other hand, Structured Reservoir Computing seems to always converge faster than Reservoir Computing. We thus confirm in the recurrent case the intriguing effectiveness of Structured Random Features \cite{choromanski2017unreasonable}, that may originate from the orthogonality of the matrix $W_r$ in SRC.

As a final remark, weight matrices in Fig.~\ref{fig:lipschitz} were not redrawn as supposed in Section~\ref{subsec: cvg theorem}. This assumption was necessary as correlations are often difficult to take into account in a theoretical setting. This is important for Reservoir Computing as it would be unrealistically slow to draw new random matrices at each time step.

\subsection{When to use RK or SRC?} \label{subsection complexity}
The two proposed alternatives to Reservoir Computing, Recurrent Kernels and Structured Reservoir Computing, are computationally efficient. To understand which algorithm to use for chaotic system prediction, we need to focus on the limiting operation in the whole pipeline of Reservoir Computing, the recurrent iterations. They correspond to Eq.~(\ref{eq: initial recurrent equation}) for RC/SRC and Eq.~(\ref{eq: recurrence k_t RI}, \ref{eq: recurrence k_t TI}) for RK. We have a time series of dimension $d$, that we split into train/test datasets of lengths $n$ and $m$ respectively. 
The exact computational and memory complexities of each step are described in Table \ref{table: computational complexity}.

\textbf{Forward:} In both Reservoir Computing and Structured Reservoir Computing, Eq. (\ref{eq: initial recurrent equation}) needs to be repeated as many times as the length of the time series. For Reservoir Computing, it requires a multiplication by a dense $N \times N$ matrix, the associated complexity scales as $O(N^2)$. On the other hand, Structured Reservoir Computing uses a succession of Hadamard and diagonal matrix multiplications, reducing the complexity per iteration to $O(N \log N)$. 

Recurrent Kernels need to recurrently compute Eq. (\ref{eq: recurrence k_t RI}, \ref{eq: recurrence k_t TI}) for all pairs of input points. For chaotic time series prediction, this corresponds to a $n \times n$ kernel matrix for training, and another kernel matrix of size  $n \times m$ for testing. To keep computation manageable, we use a well-known property in Reservoir Computing, called the Echo-State Property: the reservoir state should not depend on the initialization of the network, 
i.e. the reservoir needs to 
have a finite memory $\tau$. This property is important in Reservoir Computing and has been studied extensively \cite{jaeger2001echo,schrauwen2009computational,wainrib2016local,inubushi2017reservoir}. Transposed in the Recurrent Kernel setting, it means we can fix the number of iterations of Eq. (\ref{eq: recurrence k_t RI}, \ref{eq: recurrence k_t TI}) to $\tau$, by using a sliding window to construct shorter time series if necessary. 
A preliminary numerical study of the stability of Recurrent Kernels is presented in the Supplementary. 

\textbf{Training} requires, after a forward pass on the training dataset, to solve an $n \times N$ linear system for RC/SRC and a $n \times n$ linear system for RK. It is important to note SRC and RK do not accelerate this linear training step. We will use Ridge Regression with regularization parameter $\alpha$ to learn $W_o$. 

\textbf{Prediction} in Reservoir Computing and Structured Reservoir Computing only requires the computation of reservoir states and multiplication by the learned output weights. Recurrent Kernels need to compute a new kernel matrix for every pair $(i_r, j_q)$ with $i_r$ in the training set and $j_q$ in the testing set. Note that the prediction step includes a forward pass on the test set, followed by a linear model. 

\begin{table}[t!]
    \centering
    {\setlength{\extrarowheight}{2pt}%
    \begin{tabular}{ c|c|c|c }
    &Reservoir Computing & Structured Reservoir Computing & Recurrent Kernel
    \\[2pt]
    \hline 
    Forward & $O(nN^2)$ & $O(n N \log N)$ & $O(n^2\tau)$\\
    Training & $O(n N^2 + N^3)$ & $O(n N^2 + N^3)$ & $O(n^3)$ \\
    Prediction & $O(mN^2)$  & $O(mN\log N)$  & $O(mn\tau)$   \\[3pt]
    \hline
    Memory & $O(nN + N^2)$  & $O(nN)$  & $O(n^2 + mn)$  \\
    \end{tabular}}
    \caption{Computational and memory complexity of the three algorithms. SRC accelerates the forward pass and decreases memory complexity compared to conventional RC. The complexity of RK depends on the number of training and testing points and would be advantageous when $n\ll N$.}
    \label{table: computational complexity}
\end{table}

\begin{algorithm}[b]
\SetAlgoLined
\KwResult{Predictions $\hat{o}^{(t)} \in \mathbb{R}^{c \times m}$}
\textbf{Input:} 
A train set $\{i_r^{(t)}\}_{r=1}^n \in \mathbb{R}^{\tau \times d}$ with outputs $o \in \mathbb{R}^{c \times n}$, a test set $\{j_q^{(t)}\}_{q=1}^m \in \mathbb{R}^{\tau \times d}$.\\
 \textbf{Training:}
 Initialize an $n \times n$ kernel matrix $G^{(0)} = 0$\;
 \For{$t = 0, \ldots, \tau-1$}{
    Compute $G^{(t+1)}_{rs} = k_{t+1}(i_r^{(t)}, i_s^{(t)}, \ldots, i_r^{(0)}, i_s^{(0)})$ using Eq. (\ref{eq: recurrence k_t RI}) or (\ref{eq: recurrence k_t TI}) and $G^{(t)}_{rs}$.
 }
 Compute the output weights $W_o \in \mathbb{R}^{c \times n}$ that minimize $\|o - W_o G^{(\tau)}\|_2^2 + \alpha \|W_o\|_2^2$\;
 \textbf{Prediction:}
 Initialize an $n \times m$ kernel matrix $K^{(0)} = 0$\;
 \For{$t = 1, \ldots, \tau$}{
    Compute $K^{(t+1)}_{rq} = k_{t+1}(i_r^{(t)}, j_q^{(t)}, \ldots, i_r^{(0)}, j_q^{(0)})$ using Eq. (\ref{eq: recurrence k_t RI}) or (\ref{eq: recurrence k_t TI}) and $K^{(t)}_{rq}$.
 }
 Compute the predicted outputs $\hat{o}^{(t)} = W_o K^{(\tau)}$\;
 \caption{Recurrent Kernel algorithm}
\end{algorithm}

\section{Chaotic time series prediction}
\textbf{Chaotic time series prediction} is a task arising in many different fields such as fluid dynamics, financial or weather forecasts. By definition, it is difficult to predict their future evolution since initially small differences get amplified exponentially. Recurrent Neural Networks and in particular Reservoir Computing represent very powerful tools to solve this task \cite{antonik2018using, vlachas2019forecasting}. 

The Kuramoto-Sivashinsky (KS) chaotic system is defined by a fourth-order partial derivative equation in space and time \cite{kuramoto1978diffusion, sivashinsky1977nonlinear}. We use a discretized version from a publicly available code \cite{vlachas2019forecasting} with input dimension $d = 100$. Time is normalized by the Lyapunov exponent $\lambda = 0.043$ which defines the characteristic time of exponential divergence of a chaotic system, i.e. $|\delta x^{(t)}|\approx e^{\lambda t}|\delta x^{(0)}|$. 


KS data points $i^{(0)}, \ldots, i^{(t-1)}$ are fed to the algorithm. The output in Eq. (\ref{eq: linear output layer equation}) for Reservoir Computing consists here in predicting the next state of the system: $\hat{o}^{(t)} = i^{(t)}$. This prediction is then used for updating the reservoir state in Eq. (\ref{eq: initial recurrent equation}), the algorithm outputs the next prediction $\hat{o}^{(t+1)}$, and we repeat this operation. Thus, Reservoir Computing defines a trained \textit{autonomous dynamical system} that one wants to be synchronized with the chaotic time series \cite{antonik2018using}. 

The hyperparameters are found with a grid search, and the same set is used for RC, SRC, and RK to demonstrate their equivalence. 
To improve the performance of the final algorithm, we also add a random bias and use a concatenation of the reservoir state and the current input for prediction, replacing Eq. (\ref{eq: linear output layer equation}) by $\hat{o}^t = W_o[x^{(t)}, i^{(t)}]$. 

\begin{figure}[t!]
    \centering
    \begin{subfigure}[t]{0.69\textwidth}
        \centering
        \includegraphics[width =\linewidth]{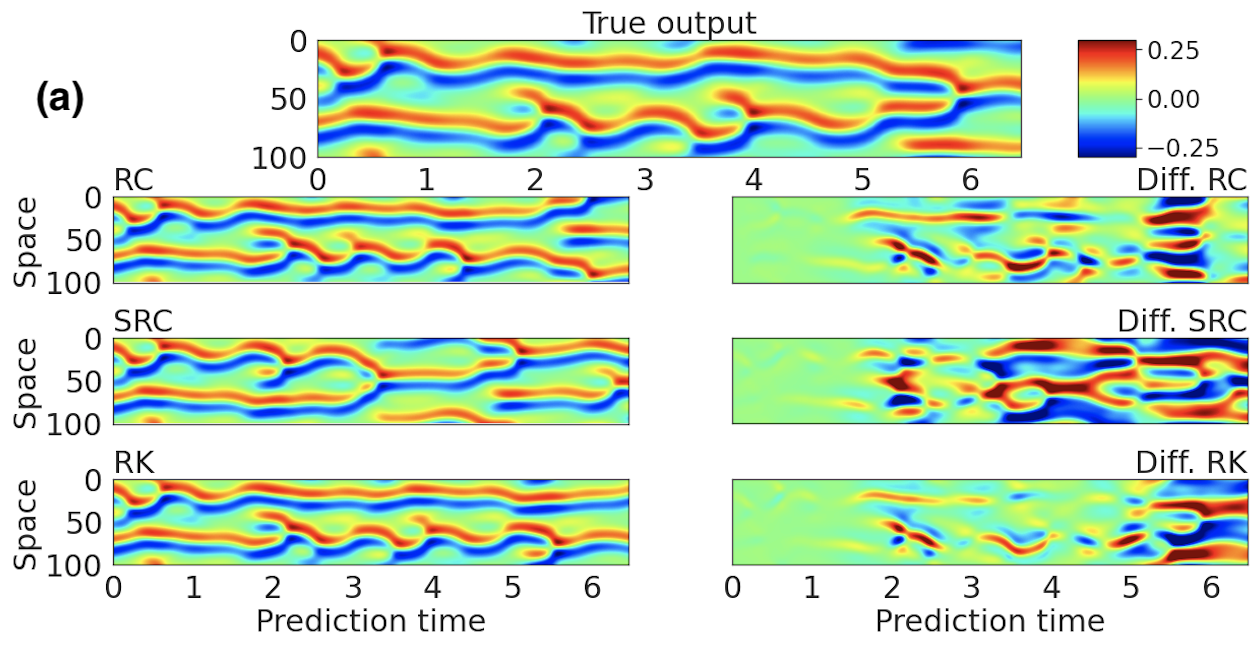}
    \end{subfigure}%
    ~ 
    \begin{subfigure}[t]{0.30\textwidth}
        \centering
        \includegraphics[width =\linewidth]{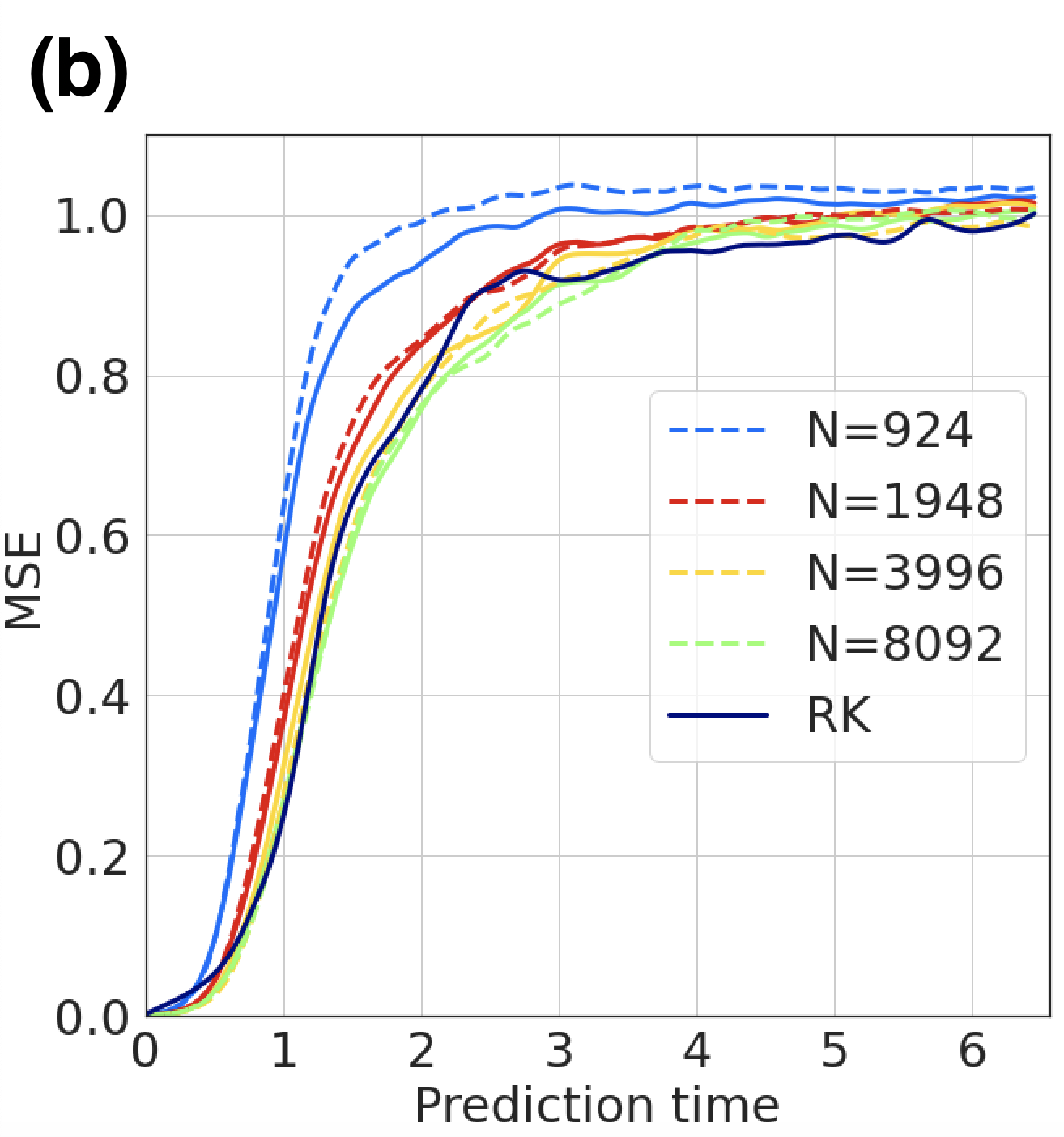}
    \end{subfigure}
    \caption{(a) Comparison of different algorithms for the prediction of the Kuramoto-Sivashinsky dataset. True output (top), predictions of RC/SRC/RK (left) and differences with the true output (right), with reservoirs in RC/SRC of size $N = 3{,}996$. We observe that each technique is able to predict up to a few characteristic times. (b) Mean-Squared Error as a function of the prediction time for RC \textbf{(full lines)}, SRC \textbf{(dashed lines)}, and RK \textbf{(black)}. For all the reservoir sizes considered, the performances of RC and SRC are very close and they converge for large dimensions to the RK limit.}
    \label{fig: ks pred}
\end{figure}

\textbf{Prediction performance} is presented in Fig.~\ref{fig: ks pred}.
RC and SRC are trained on $n=70{,}000$ training points and RK on a sub-sampling of $7{,}000$ of these training points, due to memory constraints. 
The testing dataset length was set at $2{,}000$. 
The sizes $N$ in Reservoir Computing and Structured Reservoir Computing are chosen so the dimension $p = N+d$ in Eq. (\ref{eq: structured transform}) is a power of two for the multiplication by Hadamard matrix. 
Linear regression is solved using Cholesky decomposition.

The predictions in Fig. \ref{fig: ks pred} show that all three algorithms are able to predict up to a few characteristic times. Since the prediction performance varies quite significantly between different realizations, we also display the Mean-Squared Error (MSE) of each algorithm, as a function of the prediction time and averaged over 10 realizations. We normalize each curve by the MSE between two independent KS systems.

We observe a decrease in the MSE when the size of the reservoir increases, meaning a larger reservoir yields better predictions. Performances are equivalent between RC and SRC, and they converge towards the RK performance for large reservoir sizes. Hence, this means RC, SRC, and RK can seamlessly replace one another in practical applications.

\textbf{Timing benchmark.}
Several steps in the Reservoir Computing pipeline need to be assessed separately, as described in \ref{subsection complexity}.
We present the timings on a training set of length $n=10,000$ and testing length of $m=2,000$ in Table \ref{table: timings} for all three algorithms. 

The \textit{forward pass}, i.e. computing the recurrent iterations of each algorithm, is considered separately from the linear regression for \textit{training}, to emphasize the cost of this important step.  
In RC, the most expensive operation is the dense matrix multiplication; the GPU memory was not large enough to store the square weight matrix for the two largest reservoir sizes. 
With Structured Reservoir Computing, this forward pass becomes very efficient even at large sizes, and memory is not an issue anymore. We observe that the forward pass complexity becomes approximately constant until dimension $\sim 10^5$. 
On the other hand, Recurrent Kernels iterations are very fast, as we only need to compute element-wise operations in a kernel matrix.

\textit{Prediction} requires a forward pass and then is performed with autonomous dynamics as presented on Fig.~\ref{fig: ks pred} where Eq.~\eqref{eq: linear output layer equation} is repeated $600$ times. 
For Recurrent Kernels, prediction remains slow, and this drawback is exacerbated by the autonomous dynamics strategy in time series prediction, that requires successive prediction steps. 

This shows that SRC is a very efficient way to scale-up Reservoir Computing to large sizes and reach the asymptotic limit of performance. On the other hand, the deterministic Recurrent Kernels are surprisingly fast to iterate, at the cost of a relatively slow prediction when the number of training samples $n$ is large. 

\vspace{0.2cm}
\begin{table}[h!]
    \centering
    {\setlength{\extrarowheight}{2pt}%
    \begin{tabular}{c|c|c|c|c|c}
      &$N=1{,}948$ & $N=3{,}996$ & $N=8{,}092$ & $N=16{,}284$& $N = 32{,}668$\\
    \hline
    RC & $\textbf{2.6}/0.02/1.9$& $\textbf{3.1}/0.05/4.6$&$10.4/0.16/15.4$ & Mem. Err.& Mem. Err.  \\
    SRC & $3.3/0.02/\textbf{1.6}$ & $3.4/0.05/\textbf{2.7}$ & $\textbf{3.5}/0.16/\textbf{3.7}$ &  $\textbf{3.6}/0.57/\textbf{6.8}$& $\textbf{3.6}/2.57/\textbf{13.0}$ \\
     \hline
     RK & \multicolumn{5}{c}{$\textbf{0.7}/\textbf{0.09}/23.0$}  
    \end{tabular}}
    \caption{Timing (Forward/Train/Predict, in seconds) for a KS prediction task as a function of $N$. We observe that Recurrent Kernels are surprisingly fast, except for prediction. Structured Reservoir Computing reduces drastically the speed of the forward pass at large sizes and is more memory-efficient than Reservoir Computing. Experiments were run on an NVIDIA V100 16GB.}
    \label{table: timings}
\end{table}

\section{Conclusion}

In this work, we strengthened the connection between Reservoir Computing and kernel methods based on theoretical and numerical results, and showed how efficient implementations of Recurrent Kernels can be competitive with standard RC for chaotic time series prediction. Future lines of work include a deeper study of stability and the extension to different recurrent networks topologies. We deeply think this connection between random RNNs and kernel methods will open up future research on this important topic in machine learning. 

We additionally introduced Structured Reservoir Computing, an acceleration technique of Reservoir Computing using fast Hadamard transforms. With only a simple change of the reservoir weights, we are able to speed up and reduce the memory cost of Reservoir Computing and therefore reach very large network sizes. We believe Structured Reservoir Computing offers a promising alternative to conventional Reservoir Computing, replacing it whenever large reservoir sizes are required. 


\section*{Broader Impact}

Our work consists in a theoretical and numerical study of acceleration techniques for random RNNs. Theoretical studies are important to understand machine learning to avoid relying on black boxes, towards a more responsible use of these algorithms as more and more applications appear in our daily life. 

On the other hand, efficient machine learning is necessary due to the ever-increasing power consumption required for computation. The Recurrent Kernels and Structured Reservoir Computing methods we developed pave the way towards much more efficient Reservoir Computing algorithms.

\begin{ack}
Authors would like to thank Sylvain Gigan, Antoine Boniface (Laboratoire Kastler-Brossel), and Laurent Daudet (LightOn) for interesting discussions. RO acknowledges support by grants from Région Ile-de-France. MR acknowledges funding from the Defense Advanced Research Projects Agency (DARPA) under Agreement No. HR00111890042. FK acknowledges support by the French Agence Nationale de la Recherche under grant ANR17-CE23-0023-01 PAIL and ANR-19-P3IA-0001 PRAIRIE. Additional funding is acknowledged from “Chaire de recherche sur les modèles et sciences des données”, Fondation CFM pour la Recherche-ENS.
\end{ack}


\medskip
\small
\bibliography{refs.bib}
\newpage
\pagebreak
\appendix

\section{Convergence rate for translation-invariant kernels}

\begin{theorem}\label{theorem1: TI kernels}(Rotation-invariant kernels) 
For the RI recurrent kernel defined in Eq. (\ref{eq: recurrence k_t TI}),
under the assumptions detailed above, and with $\Lambda = 2\sigma_r^2 L$ (note the factor $2$ compared to Theorem \ref{theorem1: RI kernels}). For all $t\in\mathbb{N}$, the following inequality is satisfied for any $\delta>0$ with probability at least $1-2(t+1)\delta$:
\begin{align}
    \label{eq: main theorem bound TI}
    \left|\langle x^{(t+1)},y^{(t+1)}\rangle  - k_{t+1}(\Delta^{(t)},...,\Delta^{(0)})\right| 
    &\leq \frac{1-\Lambda^{t+1}}{1-\Lambda} \Theta(N) &&\textrm{if} \quad \Lambda\neq 1 \\
    \label{eq: main theorem bound 2 TI}
    &\leq (t+1) \Theta(N) &&\textrm{if} \quad \Lambda=1
\end{align}
with $\Theta(N) = \frac{4\kappa^2\log\frac1\delta}{3N}+2\kappa^2\sqrt{\frac{2\log\frac1\delta}{N}}$.
\end{theorem}
\vspace{-0.2cm} 
\begin{proof}

Under the assumptions, Proposition~\ref{Prop1: Bernstein} yields with probability greater than $1-2\delta$:
\begin{equation}
    \label{eq: bernstein scalar product }
    \left|\langle x^{(t+1)},y^{(t+1)}\rangle -k([x^{(t)}, i^{(t)}],[y^{(t)}, j^{(t)}])\right| \leq \frac{4\kappa^2\log\frac1\delta}{3N}+2\kappa^2\sqrt{\frac{2\log\frac1\delta}{N }} = \Theta(N)
\end{equation}
It means the larger the reservoir, the more Random Features $N$ we sample, and the more the inner product of reservoir states concentrates towards its expectation value, at a rate $O(1/\sqrt{N})$. We now apply this inequality recursively to complete the proof, based on the observation that both Eq. (\ref{eq: main theorem bound}) and (\ref{eq: main theorem bound 2}) are equivalent to: $\left|\langle x^{(t+1)},y^{(t+1)}\rangle  - k_{t+1}(\Delta^{(t)},...,\Delta^{(0)})\right| \leq (1 + \Lambda + \Lambda^2 +... + \Lambda^t)\Theta(N)$.

For $t = 0$, provided $x^{(0)} = y^{(0)} = 0$, we have, according to Eq.~\ref{eq: bernstein scalar product}, with probability at least $1 - 2 \delta$:
\begin{equation}
    \left|\langle x^{(1)}, y^{(1)}\rangle - k_1(\Delta^{(0)})\right|\leq \Theta(N)
\end{equation}

For any time $t \in \mathbb{N}^*$, let us assume the following event $A_{t}$ is true with probability $\mathbb{P}(A_{t}) \geq 1 - 2 t \delta$:
\begin{equation}
    \left|\langle x^{(t)}, y^{(t)}\rangle - k_{t} (\Delta^{(t-1)},...,\Delta^{(0)}) \right|
    \leq (1 + \ldots + \Lambda^{t-1}) \Theta(N)
\end{equation}
Using the Lipschitz-continuity of $k$, this inequality is equivalent to:
\begin{equation}
    \label{eq: proof final eq to t TI}
    \left|
    k(2\sigma_r^2(1- \langle x^{(t)}, y^{(t)}\rangle) + \Delta^{(t)})
    - k(2\sigma_r^2(1- k_{t} (\Delta^{(t-1)},...,\Delta^{(0)})) + \Delta^{(t)}) \right|
    \leq (\Lambda + \ldots + \Lambda^{t}) \Theta(N)
\end{equation}

With Eq. (\ref{eq: bernstein scalar product}), the following event $B_{t}$ is true with probability $\mathbb{P}(B_{t}) \geq 1 - 2 \delta$:
\begin{equation}
\label{eq: proof eq t+1 TI}
\bigg|\langle x^{(t+1)}, y^{(t+1)}\rangle- k\big(2\sigma_r^2(1-\langle x^{(t)}, y^{(t)}\rangle)+\Delta^{(t)}\big)\bigg|\leq \Theta(N)    
\end{equation}

Summing Eq. (\ref{eq: proof final eq to t TI}) and (\ref{eq: proof eq t+1 TI}), with the triangular inequality and a union bound, the following event $A_{t+1}$ is true with probability $\mathbb{P}(A_{t+1}) \geq \mathbb{P}(B_{t} \cap A_t) = \mathbb{P}(B_{t})+\mathbb{P}(A_{t})-\mathbb{P}(B_{t}\cup A_{t})\geq 1-2\delta + 1-2t\delta - 1 = 1 - 2 (t+1) \delta$:
\begin{equation}
    \left|\langle x^{(t+1)}, y^{(t+1)}\rangle
    -
    k_{t+1} (\Delta^{(t)},...,\Delta^{(0)})\right|
    \leq (1 + \ldots + \Lambda^{t}) \Theta(N)
\end{equation}
\end{proof}

\section{Explicit Recurrent Kernel formulas}
\label{sec: explicit rk}

We have defined so far the general formulas of RI and TI Recurrent Kernels in Eq. (\ref{eq: recurrence k_t RI}) and (\ref{eq: recurrence k_t TI}). We will give now their explicit formulas for specific activation functions that one may encounter in Reservoir Computing.

Two reservoirs $x^{(t)}$ and $y^{(t)}$ are driven by two respective input time series $i^{(t)}$ and $j^{(t)}$. They obey Eq. (\ref{eq: initial recurrent equation}) and in the infinite-size limit, their inner product converges towards an explicit Recurrent Kernel. In practice, one needs to compute the inner products for each pair of input time series, from the training or testing sets, that we concatenate to construct a kernel matrix. 

A list of different activation functions and their associated kernels is provided in Table \ref{Table: RF and kernels}. Without recurrence, it is always possible to write the corresponding kernel as an integral that one may evaluate:
\begin{equation}
    \label{eq: k(u,v) as an integral}
    k(u,v) = \int dw \rho(w) 
        f( \langle w, u \rangle ) f( \langle w, v \rangle )
\end{equation}
where $\rho(w)$ is the distribution of the weights, usually an i.i.d. gaussian distribution. However, in all the cases presented here, $k(u,v)$ happens to contain inner products $\langle u, v \rangle$, which makes it possible to define the corresponding Recurrent Kernel. 

In our case, $u^{(t)} = [\sigma_r x^{(t)}, \sigma_i i^{(t)}]$ and $v^{(t)} = [\sigma_r y^{(t)}, \sigma_i j^{(t)}]$ so that:
\begin{equation}
    \langle u^{(t)}, v^{(t)} \rangle = \sigma_r^2 \langle x^{(t)}, y^{(t)} \rangle + \sigma_i^2 \langle i^{(t)}, j^{(t)} \rangle
    \rightarrow \sigma_r^2 k_t(l^{(t-1)}, \ldots, l^{(0)}) + l^{(t)}
    \label{eq: inner product u v}
\end{equation}
when the reservoir size $N \rightarrow \infty$. Similarly, $\|u^{(t)}\|^2 = \langle u^{(t)}, u^{(t)}\rangle$ and $\|v^{(t)}\|^2$ are symmetric inner products that can similarly be expressed as in Eq. (\ref{eq: inner product u v}). Hence, the Recurrent Kernel formulas are derived from the previous one by noting that:
\begin{equation}
    \lim_{N\rightarrow\infty} \langle x^{(t+1)}, y^{(t+1)} \rangle 
    =
    k_{t+1} (l^{(t)}, \ldots, l^{(0)})
    \equiv
    k(u^{(t)}, v^{(t)})
\end{equation}

Analytic formulas in more general cases may not exist and they would need to be replaced by successive integrals. In this work, we restricted ourselves to functions described in Table 1 with simple analytic formulas, to speed up the RK computation. For instance, the error function is very close but not equal to the hyperbolic tangent in our implementations of Reservoir Computing, and performance in practice is very similar.

The successive integrals can still be explicitly defined. Eq. (\ref{eq: k(u,v) as an integral}) describes the asymptotic kernel limit for any arbitrary $(u,v)$. To define recurrent kernels, we need to express it as a function of $\langle u, v \rangle$, $\|u\|^2$, and $\|v\|^2$ only. This is possible thanks to the invariance by rotation of the gaussian distribution of $w$. Without loss of generality, we can thus assume that $u = \|u\| e_1$ and $v = \|v\| (\cos\theta e_1 + \sin\theta e_2)$ with $e_1$ and $e_2$ the first two vectors of the canonical basis and $\theta = \langle u, v \rangle / (\|u\| \|v\|)$ (which is a function of the three quantities of interest). The multidimensional integral boils down to a two dimensional integral:
\begin{equation}
    k(u,v) = \int\int dw_1 dw_2 \rho(w_1) \rho(w_2) f(w_1 \|u\|) f(\|v\| (w_1\cos\theta + w_2\sin\theta) )
\end{equation}
where $w_1$ and $w_2$ are gaussian random variables, projections of $w$ on $e_1$ and $e_2$. Hence it is possible to iterate Recurrent Kernels numerically, that are the large-size limit of any Reservoir Computing algorithm for every activation function $f$. Each component of the square kernel matrix would require the evaluation of this two-dimensional integral, it may be possible to use tabular values to speed up computation. 

\begin{table}[t]
\centering
{\setlength{\extrarowheight}{0.4cm}
\begin{tabular}{c|c} 
 $f(\cdot)$ & Associated kernel $k(u,v)$ \\[5pt] 
 \hline 
 Erf($\cdot$) & $\frac{2}{\pi} \arcsin\left(\frac{2\langle u,v\rangle}{\sqrt{(1+2\|u\|^2)(1+2\|v\|^2)}}\right)$ \\
 RFFs: $[\cos(\cdot), \sin(\cdot)]$ & $\exp\left(-\frac{\|u-v\|^2}{2}\right)= \exp\left(-\frac{\|u\|^2+\|v\|^2-2\langle u,v\rangle}{2}\right)$ \\
 Sign($\cdot$) & $\frac{2}{\pi}\arcsin\left(\frac{\langle u,v\rangle}{\|u\|\|v\|}\right)$  \\
 Heaviside($\cdot$) & $\frac12 - \frac{1}{2\pi}\arccos\left(\frac{\langle u,v\rangle}{\|u\|\|v\|}\right)$\\
 ReLU($\cdot$) & $\frac{1}{2\pi} \left(\langle u,v\rangle\arccos(-\frac{\langle u,v\rangle}{\|u\|\|v\|})+\|u\|\|v\|\sqrt{1-\left(\frac{\langle u,v\rangle}{\|u\|\|v\|}\right)^2}\right)$  \\ 
\end{tabular}}
 \caption{Table of point-wise non-linearities $f$ and their approximated kernels. For any $u, v \in \mathbb{R}^p$ the kernel $k(u,v)$ is the limit when $N$ goes to infinity of $\frac{1}{N}\langle f(Wu),f(Wv)\rangle$ with $W \in \mathbb{R}^{N \times p}$ an i.i.d. normal random matrix. In the case of Reservoir Computing, we have $u = u^{(t)} = [\sigma_r x^{(t)}, \sigma_i i^{(t)} ]$ and $v = v^{(t)} = [\sigma_r y^{(t)}, \sigma_i j^{(t)}]$. We observe that in this table, all kernel formulas depend only on $\langle u, v \rangle$, $\|u\|$, and $\|v\|$, which makes it possible to easily derive the Recurrent Kernel equations. }
 \label{Table: RF and kernels}
\end{table}

\section{Numerical study of the independence hypothesis}

One assumption for the previous convergence theorems states the weight matrices $W_r$ and $W_i$ have to be redrawn at each iteration. This independence hypothesis is required in Eq. (\ref{eq: proof eq t+1}) and Eq. (\ref{eq: proof eq t+1 TI}), to ensure that $x^{(t)}$ and $y^{(t)}$ are uncorrelated with the weight matrices. This is necessary in the theoretical study to properly define the expectations and ensure the i.i.d. requirement for the random variables in the Bernstein inequality. 

However, this assumption is unrealistic for practical Reservoir Computing. Resampling weight matrices at each timestep is computationally demanding and output weights would depend on the realization of these random matrices: one would need to keep the same random matrices in memory for testing.

\begin{figure}
    \centering
    \includegraphics[width=\linewidth]{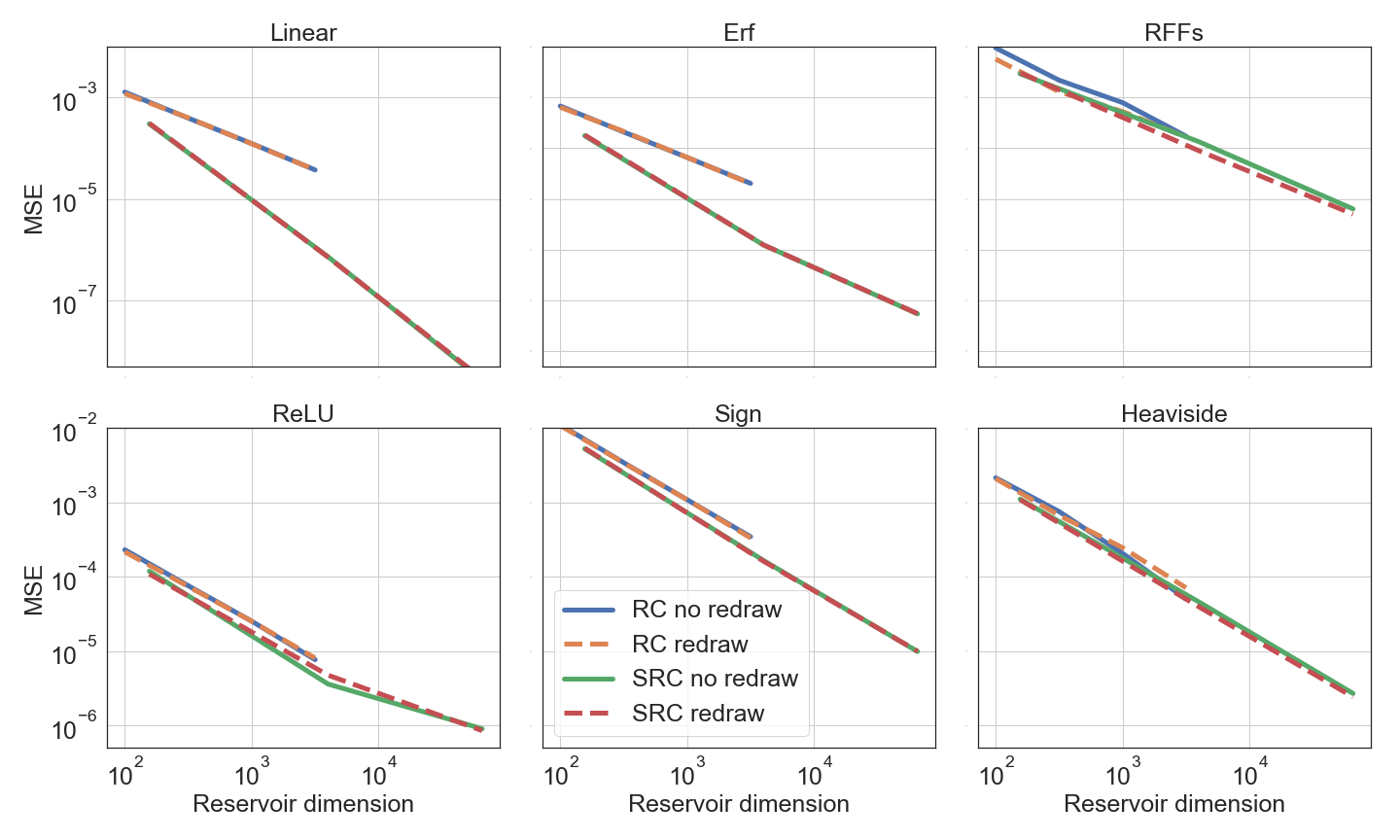}
    \caption{Mean-Squared error between the kernel matrix obtained with RC/SRC with the asymptotic kernel limit, with and without resampling the random matrices at each iteration, to test the independence hypothesis of the theorem. $50 \times 50$ kernel matrices have been generated for all pairs of 50 random input time series of length 10. Several activation functions and their corresponding recurrent kernels are presented here. We observe that the hypothesis does not seem to be necessary since RC and SRC without resampling also converge to the RK limit at sensibly the same speed.}
    \label{fig:redraw}
\end{figure}

However, in Fig. \ref{fig:redraw}, we investigate the convergence with and without redrawing weights at each iteration, and this independence hypothesis does not seem to be necessary: convergence is still achieved with fixed weight matrices. We show the Mean-Squared Error $\|K_1 - K_2\|_2^2 / n^2$ between the kernel matrix $K_1$ from the explicit RK formula and $\hat{K}_2$ the one obtained with RC and SRC, with and without redrawing the random matrices at every timestep. Each kernel matrix is of size $50 \times 50$, as we use $n = 50$ random i.i.d gaussian input time series of dimension $50$ and time length $10$. Each curve is an average over $10$ realizations and the reservoir scale is set to $\sigma_r^2 = 0.25$ to ensure stability. 

We confirm the observation from Fig. \ref{fig:lipschitz} that the larger the reservoir dimension, the closer we are from the RK asymptotic limit. This is valid for several activation functions, the ones presented in Table \ref{Table: RF and kernels}. We also confirm that SRC generally converges faster than RC. 

Convergence is still achieved when resampling the weights at each iteration, and speed of convergence is not significantly different than for the fixed random matrix case. Thus convergence seems to be much more robust in practice, and this may call for further theoretical studies. 

\section{Stability of Reservoir Computing and Recurrent Kernels}

As the reservoir is itself a dynamical system, it can be stable (differences in initial conditions vanish with time) or chaotic (differences in initial conditions explode exponentially). This is linked with the Echo-State Property, extensively studied for Reservoir Computing. It states that two reservoirs initialized differently need to converge to the same trajectory, provided they share the same weights (at each time step if weights are resampled). This property is important so that the reservoir state after a large enough time $\tau$ does not depend on the arbitrary reservoir initialization. Stability or chaos can be tuned depending on a set of hyperparameters. An important one is the scale of the reservoir weights: when small, initial differences get damped exponentially with time, whereas they may explode if reservoir weights are large. 

\begin{figure}
    \centering
    \includegraphics[width=\linewidth]{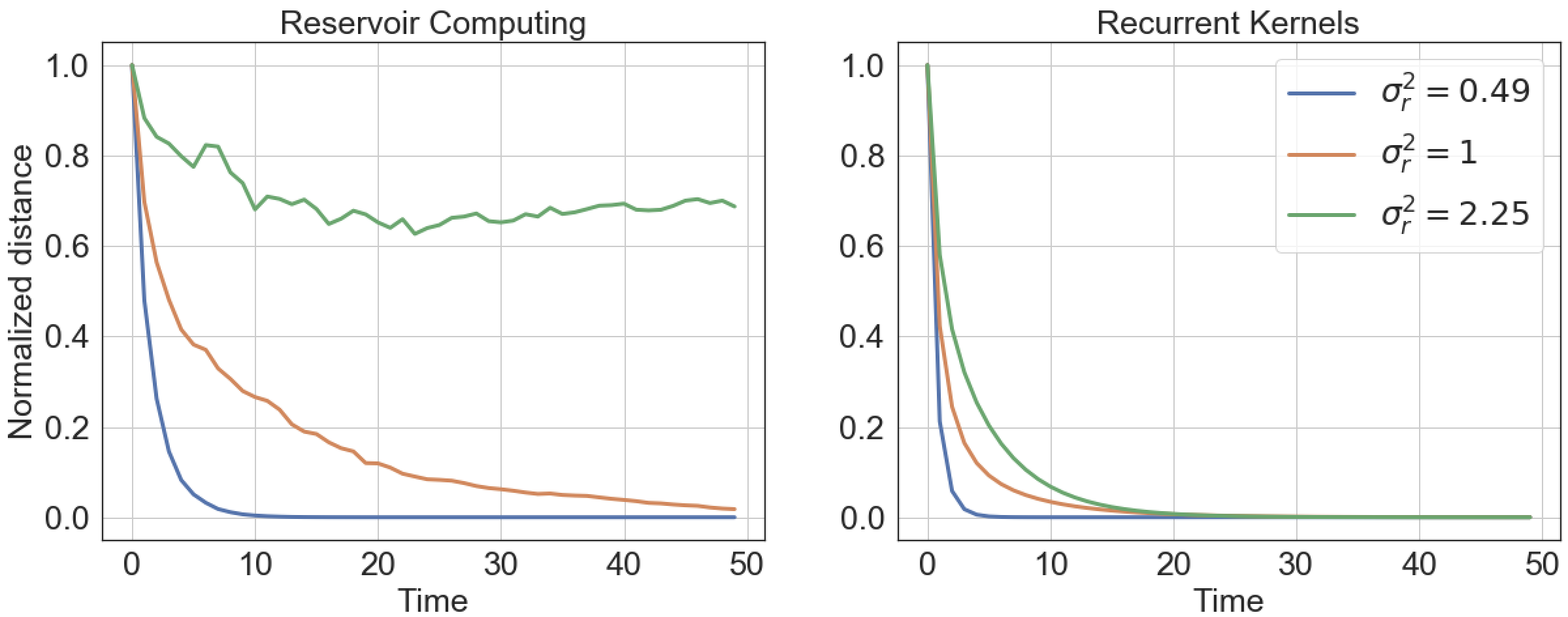}
    \caption{Stability analysis of Reservoir Computing and Recurrent Kernels. We compute the normalized square distance between two reservoirs or recurrent kernels fed with the same input time series and different initializations. For RK or RC when $\sigma_r^2 \leq 1$ we see that trajectories converge to a single one after some time. This means that initial conditions are forgotten after a number of iterations. On the other hand, when $\sigma_r^2 = 2.25$ for Reservoir Computing, the reservoir is in a chaotic regime and always depend on initial conditions. It is interesting to observe that Recurrent Kernels are generally more stable than RC.}
    \label{fig:stability}
\end{figure}

We verify this Echo-State Property here for Reservoir Computing. In Fig. \ref{fig:stability} we present the squared distance $\|x_1^{(t)} - x_2^{(t)}\|^2$ as a function of time $t$ between two randomly initialized reservoirs $x_1$ and $x_2$, for the same input time series from the Kuramoto-Sivashinsky dataset. A normalization factor has been added to normalize this distance to 1 at $t=0$ and each curve is an average over $100$ realizations. The activation is the error function, the input scale is set to a small value $\sigma_i^2 = 0.01$, and we vary the reservoir scale $\sigma_r^2$. For $\sigma_r^2 = 0.49$ and $1$, dynamics are stable and the two reservoir states converge quite quickly to the same trajectory. 
When $\sigma_r^2 = 2.25$, dynamics becomes chaotic and the two reservoirs follow very different dynamics due to their different initial conditions.

Recurrent Kernels may also present this transition from stability to chaos. Moreover, this stability property is important for Recurrent Kernels in practice. RKs need to be iterated a certain number of times, and thanks to stability this number of iterations can be reduced to the finite memory $\tau$ and not on the full length of the time series. This change reduces considerably the computational costs.

We thus also investigate numerically the stability of Recurrent Kernels, i.e. how they depend on the initial conditions. In Fig. \ref{fig:stability}, we present the normalized difference between two kernel matrices $\|K_1^{(t)} - K_2^{(t)}\|_2^2$ as a function of time, for two recurrent kernels $K_1$ and $K_2$ initialized with a matrix full of ones or of zeros, and fed with the same input time series, for the arcsine Recurrent Kernel corresponding to the erf activation function. We observe that Recurrent Kernels are in general a lot more stable than Reservoir Computing. This characteristic may be interesting to investigate further. 

We may now draw an interesting parallel between this study and, as we unroll the Recurrent Neural Network through time, multilayer perceptrons with random weights, linked with compositional kernels. They correspond to our case, $i^{(t)} = 0$ for $t\geq1$ and $i^{(0)} \in \mathbb{R}^d$ is the time-independent input. This stability property corresponds to a final layer that does not depend on $i^{(0)}$, and as such information does not flow in the deep network. Hence, whereas it is advantageous in Reservoir Computing to be stable, it may be detrimental for deep neural networks. 

\section{Implementation details for Reservoir Computing}

Several tweaks are useful to improve the performance of Reservoir Computing for time series prediction. We used the erf activation function as it is the closest from the hyperbolic tangent already used in Reservoir Computing, that still possess a simple Recurrent Kernel formula. 

First, we add a random additive bias $b \in \mathbb{R}^N$ sampled from an i.i.d. normal distribution $\mathcal{N}(0, \sigma_b^2)$. The variance of this bias vector $\sigma_b^2$ is a hyperparameter to tune, like the variance of the reservoir or input weights. This bias helps to diversify the neuron activations in the reservoir. Hence, the reservoir update equation becomes:
\begin{equation}
	\label{eq: recurrent equation with bias}
	x^{(t+1)} = \frac{1}{\sqrt N} f\left(W_r\,x^{(t)} + W_i\,i^{(t)} + b\right)
\end{equation}

As stated previously, we concatenate the reservoir state with the last value of the time series we have received. Information about the past is still encoded in the reservoir, but with this simple change, the reservoir is rather used to compute perturbations on the current value, and does not have to reconstruct the whole spatial profile. We add a renormalization hyperparameter $r$ for this concatenation, in order to control the weight of the reservoir versus current input.

A hyperparameter search was performed, for a total of 5 hyperparameters (the reservoir scale, input scale, bias scale, the previous concatenation factor, regularization constant). Since there is a large number of hyperparameters to tune, we perform it on one hyperparameter at a time, going through the set of parameters several times. The final set of hyperparameters of Fig. \ref{fig: ks pred} is $\{\sigma_i, \sigma_r, \sigma_b, r, \alpha\} = \{0.4, 0.9, 0.4, 1.1, 10^{-2}\}$.

For completeness, we give here the exact definition of the Mean-Squared Error of Fig. \ref{fig: ks pred}. The target output $O(t) \in \mathbb{R}^{d}$ for $t = 1, \ldots, T_{\rm{pred}}$ corresponds to the next states of the chaotic systems, and for each $t$, we evaluate the MSE between $O(t)$ and the prediction of the algorithm $\hat{O}(t)$, which is simply $\|O(t) - \hat{O}(t)\|^2 / d$.

\section{Implementation details for Recurrent Kernels}

We also used a Recurrent Kernel to perform chaotic time series prediction. We chose an arcsine rotation-invariant kernel, the asymptotic limit of a reservoir with error function activations. We use the principle described in Section \ref{sec: explicit rk}, with the addition of a random gaussian bias that corresponds to adding a constant dimension to the vector $u^{(t)} = [\sigma_r x^{(t)}, \sigma_i i^{(t)}, \sigma_b]$.

Additionally, we have introduced for Reservoir Computing a concatenation step we need to reproduce with Recurrent Kernels. In RC, we concatenate the reservoir and the current input before computing the prediction. The corresponding operation for Recurrent Kernels is the addition of a linear kernel computed from all pairs of input points: $K^+_{kl} = \langle i_k^{(t)}, i_l^{(t)} \rangle$. This kernel matrix $K^+$ is added to the Recurrent Kernel after the iterations and before the linear model for prediction. 

\begin{figure}[t]
    \centering
    \includegraphics[width=\linewidth]{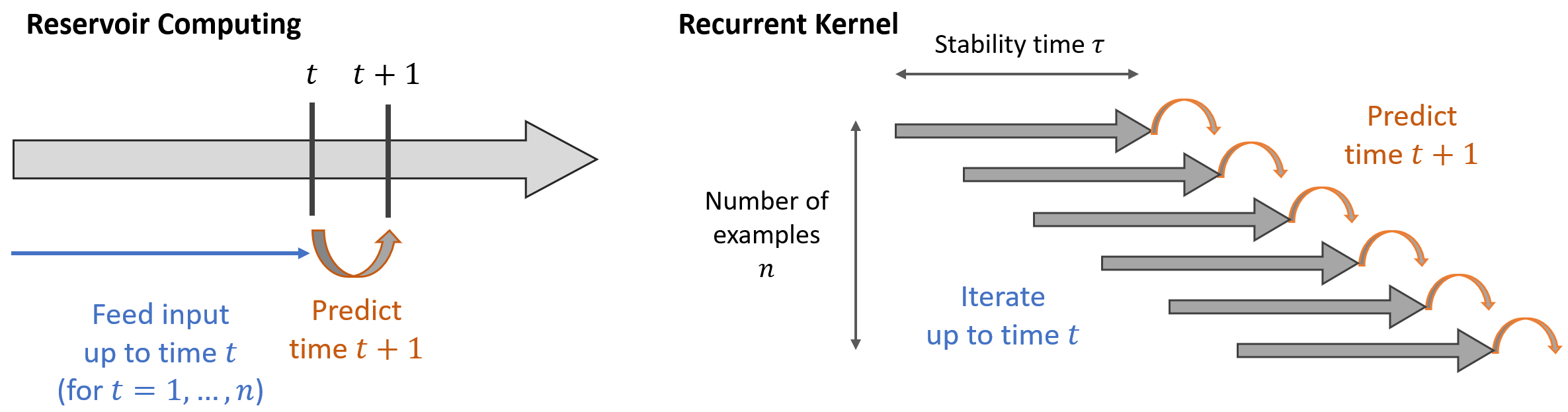}
    \caption{How to use Recurrent Kernels for time series prediction. In Reservoir Computing, the input is continuously fed to the reservoir and all the reservoir states for every timestep $t$ are stored for training. With Recurrent Kernels, we construct $n$ small windows of the time series of length $\tau$ and compute scalar products between each pair using $\tau$ iteration of Eq. (\ref{eq: recurrence k_t RI}) or (\ref{eq: recurrence k_t TI}).}
    \label{fig:explanation_rcrk}
\end{figure}

We also expand more on the process of generating the input data for Recurrent Kernels. In time series prediction, each reservoir state (neglecting a warm-up phase) is used during training to learn output weights to predict the future states of the system. Since there are $n$ training examples, this corresponds to an $n \times n$ kernel matrix. In the Recurrent Kernel setting, we train a linear model on the final kernel matrix. We thus construct $n$ time series of length $\tau = 50$ for each time step of the training data (neglecting the effect of edges), where the length $\tau$ is determined by the stability of the Recurrent Kernel. This process is depicted in Fig. \ref{fig:explanation_rcrk}.

\section{Recursive vs non-recursive prediction}

Following previous strategies developed for chaotic time series prediction with Reservoir, RC, SRC, and RK algorithms were trained only to perform next-time-step prediction. To predict further in the future, this prediction is then fed back into the algorithm to iterate further in time. As explained previously, this defines an autonomous dynamical system that should be synchronized with the chaotic time series if training is successful. 

Another possible strategy would be to use a given reservoir state to predict $T_{\rm{pred}}$ time steps in the future. The output dimension $c = d\, T_{\rm{pred}}$ is larger and the learning task becomes more difficult. 

\begin{figure}[ht!]
    \includegraphics[width=\linewidth]{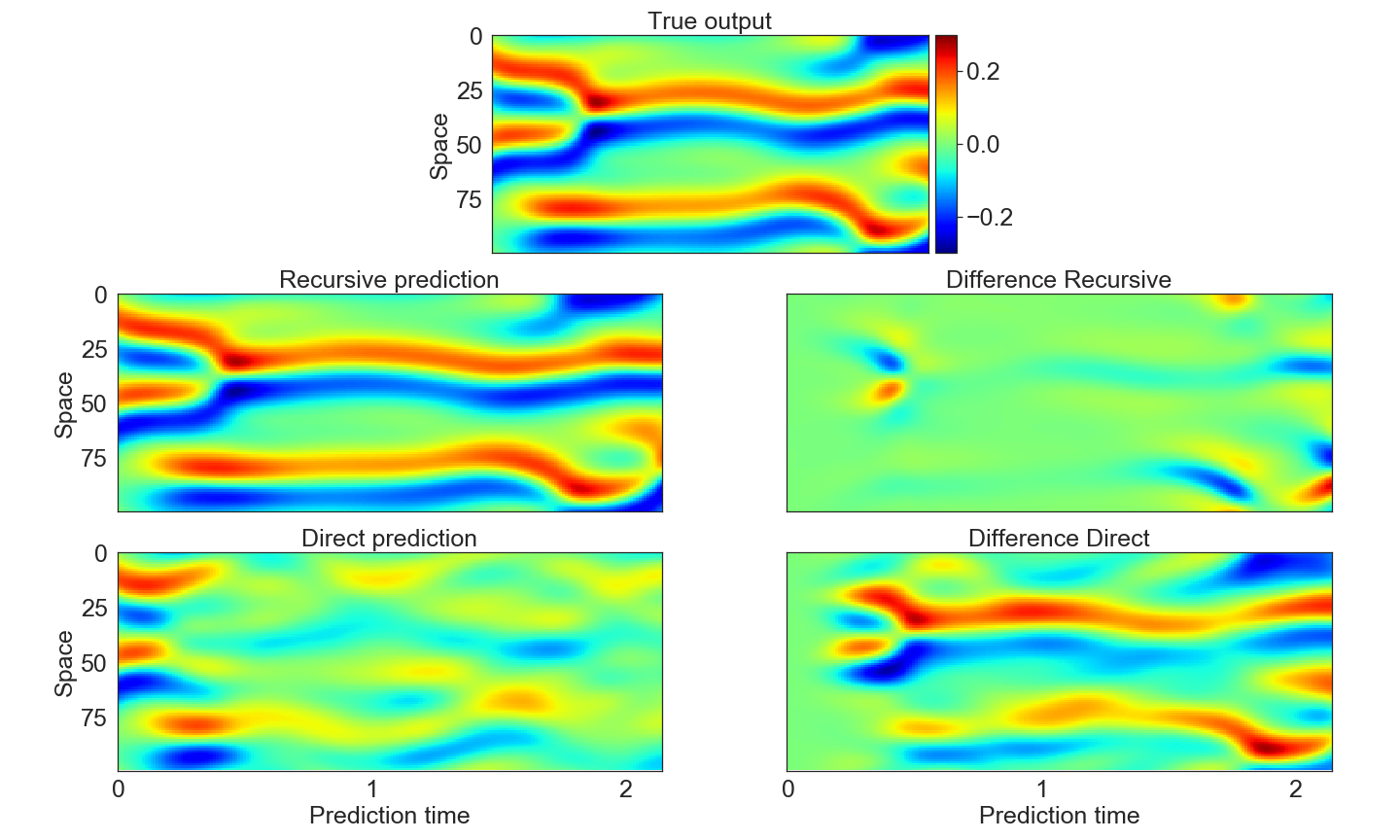}
    \caption{Comparison of recursive and non-recursive prediction. We see that with recursive prediction (left), Reservoir Computing is able to predict quite precisely up to at least 2 characteristic times. On the other hand, without recursive prediction, Reservoir Computing quickly has a hard time to guess the future of the KS system and outputs its mean for long prediction times. }
    \label{fig:no_rec}
\end{figure}

We show here the usefulness of this strategy based on autonomous dynamics. In Fig. \ref{fig:no_rec}, we show the performance of Reservoir Computing prediction on the Kuramoto-Sivashinsky dataset, with and without recursive prediction. With recursive prediction (left), this corresponds to the strategy already presented in Fig. \ref{fig: ks pred}, and it is not surprising that prediction up to at least 2 Lyapunov exponents is possible. Without recursive prediction (right), the algorithm has a much harder time to predict the future of the chaotic system. Instead, after a short while, it only returns the average value of the time series.

Note that the same hyperparameters were used in both cases. While it may be possible to improve the performance of the direct prediction strategy, by increasing the size of the reservoir or playing with regularization parameter, but we show here the simplicity and effectiveness of the recursive prediction strategy.
\newpage

\end{document}